\documentclass[preprint,12pt,review]{elsarticle}

\usepackage{amssymb}
\usepackage{epsfig}
\usepackage{graphicx}
\usepackage{amsmath}
\usepackage{amssymb}
\usepackage{array}
\usepackage{color}
\usepackage{bm}

\usepackage{multirow}
\usepackage{booktabs}
\usepackage{blkarray}
\usepackage{pmat}
\usepackage{lineno} 

\usepackage[breaklinks=true,breaklinks=true,letterpaper=true,colorlinks,bookmarks=false]{hyperref}

\journal{Pattern Recognition}

\begin{document}

\begin{frontmatter}

\title{Lane Detection with Versatile AtrousFormer and Local Semantic Guidance}

\author[label1]{Jiaxing Yang}
\address[label1]{School of Information and Communication Engineering,
                 Dalian University of Technology, Dalian, 116023, China}



\author[label1]{Lihe Zhang\corref{cor1}}
\cortext[cor1]{Corresponding author}
\cortext[cor1]{}

\author[label1]{Huchuan Lu}

\begin{abstract}
 Lane detection is one of the core functions in autonomous driving and has aroused widespread attention recently. The networks to segment lane instances, especially with bad appearance, must be able to explore lane distribution properties. Most existing methods tend to resort to CNN-based techniques. A few have a try on incorporating the recent adorable, the seq2seq Transformer~\cite{transformer}. However, their innate drawbacks of weak global information collection ability and exorbitant computation overhead prohibit a wide range of the further applications. In this work, we propose Atrous Transformer (AtrousFormer) to solve the problem. Its variant local AtrousFormer is interleaved into feature extractor to enhance extraction. Their collecting information first by rows and then by columns in a dedicated manner finally equips our network with stronger information gleaning ability and better computation efficiency. To further improve the performance, we also propose a local semantic guided decoder to delineate the identities and shapes of lanes more accurately, in which the predicted Gaussian map of the starting point of each lane serves to guide the process. Extensive results on three challenging benchmarks (CULane, TuSimple, and BDD100K) show that our network performs favorably against the state of the arts.
\end{abstract}


\end{frontmatter}


\section{Introduction}

\label{sec:intro}

\begin{figure}[tbp]
	\centering
	\includegraphics[scale=0.5]{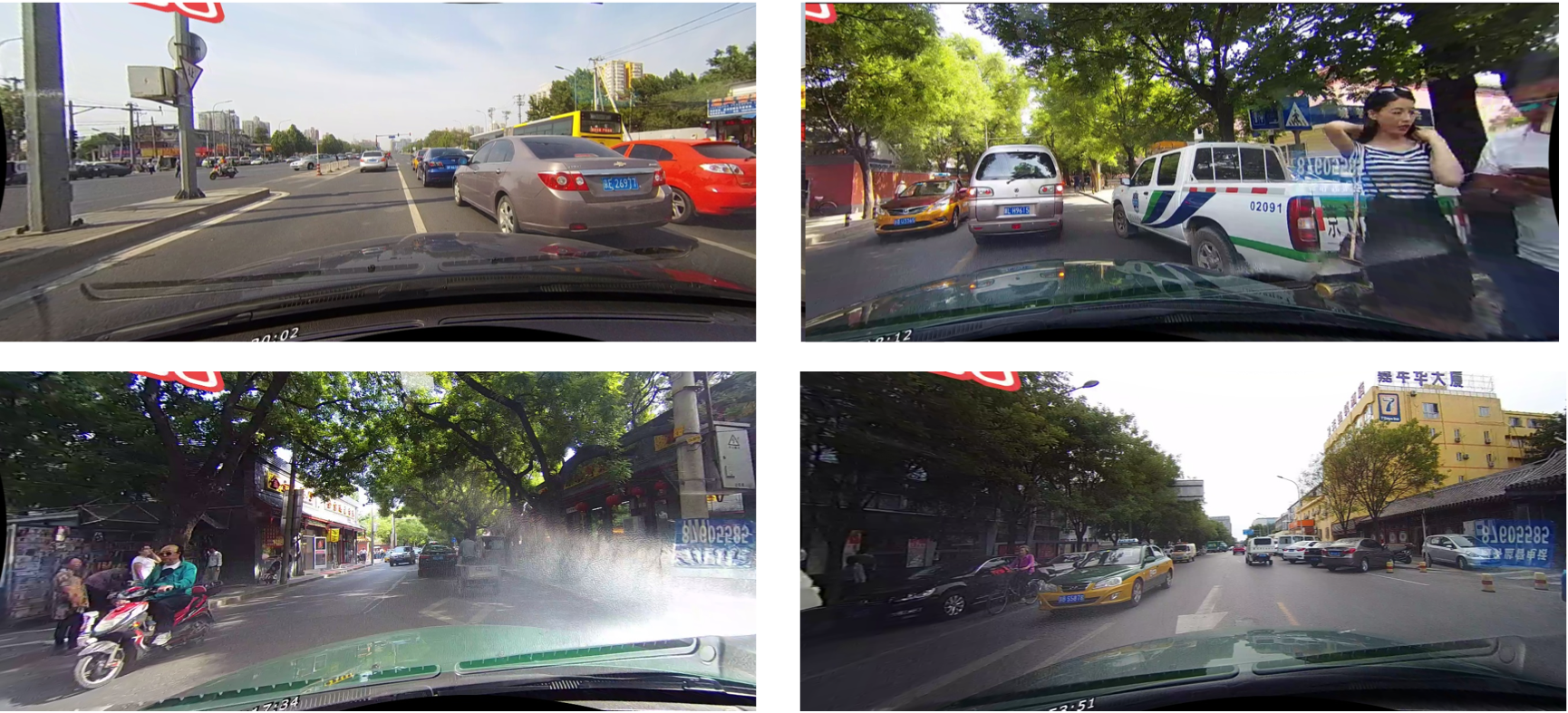}
	\caption{Visualization of lanes in bad cases: Occlusion, Crowded, High Light, and Arrow Shape.}
	\label{fig:cases}
\end{figure}

Autonomous driving becomes increasingly promising recently. In such a systematic engineering project, one core desirement is to generate high-quality lane instances (both online and offline). The vehicles rely on the generated results to keep itself from overstepping the boundaries~\cite{lanekeeping}. However, lane detection faces big challenges from the wild scenes (see Fig.~\ref{fig:cases}). Lanes may be ambiguous due to occlusion by the vehicles, bad weather conditions, abrasion, illumination intensity, and etc. In addition, the task also suffers from the disconnected, curved and slender shapes of lane itself. To solve the problems, traditional methods~\cite{traditional1,traditional2,traditional3, traditional4,traditional5,traditional6,traditional7,traditional8} tend to adopt a two-stage strategy. They first use hand-crafted operators to extract the feature, and then apply an adjustment step in order to accommodate the real line shape, such as Hough transform~\cite{traditional1,traditional2} and Random Sampling~\cite{traditional4,traditional5}. Nevertheless, although some progresses have been made, these methods are not good enough to handle complex traffic scenes. 

Recently, as deep learning based techniques gradually dominates various computer vision tasks, community begins to focus on the CNN-based techniques to facilitate the research of lane detection. Most methods define lane detection as a pixel-level semantic segmentation problem~\cite{rbnet, erfnet, liulane, SCNN, distillation, vanishing, gan, resa}. Specifically, in these methods, each lane is regarded as a semantic class. They really rely on a strong feature to infer the ideal shapes. Another group of them draw on the recent achievements in object detection. They~\cite{linecnn,laneatt,polylanenet,lstr} are developed towards two directions: anchor-based~\cite{linecnn,laneatt} and anchor-free~\cite{polylanenet,lstr}. The anchor-based ones use beaming lines starting from either left, right, or bottom as anchor baseline, and then adjust offset to fit the real shape of lane line via a post-processing step. They heavily rely on pre-defined straight lines, thus struggling to handle more complex line shapes. As for those anchor-free, they equate lane detection to high-order polynomial regression, straightforward yet overly relying on certain parameters. The last main group~\cite{key1,PINet,laneaf,curvelane}, inspired by the fancy thoughts from human pose estimation, usually extract key points with lane semantic and then cluster them into different lane instances via complex post-processing methods. In general, methods other than sementic segmentation have difficulties in modeling more complex lane forms, like those described in BDD100K~\cite{bdd}. 

In this paper, we walk further along the way of semantic segmentation based lane detection. Hereafter we get a closer look on its members. Early of them such as \cite{erfnet,rbnet,gan,liulane} have a CNN-based encoder, a plain upsampling decoder, and a binary classifier. However, their quantitative and visualization results indicate that the segmented lanes are dragged down by low-quality feature map. The problem is generated due to their ignorance towards the spatial cues of lanes and incapability of conjecturing according to the environment information. To overcome, recent explorers such as SCNN \cite{
SCNN} and RESA~\cite{resa} propagate information between near or remote feature slices in four directions (upwards, downwards, rightwards, and leftwards). However, recurrent reinforcement manner does not treat features equally and therefore does not perform well (similar to LSTM). More advanced way still needs exploring. Another problem haunting the prediction of lane instances is their weak discrimination ability, that is to say, the classifiers feeding on global representation blunder in predicting the existence of lanes.

Instead of focusing on the temporally-recurrent CNN-based techniques, in this paper, we resort to designing Atrous Transformer (AtrousFormer). In the same spirit as ASPP~\cite{aspp}, the AtrousFormer instills atrous experience to the transformer structure~\cite{transformer} by following a two stage strategy. It collects information first along rows and then along columns. Compared to ~\cite{transformer}, our designing can save lots of computation cost (has reduced the number of key\&value entities) and improve performance, simultaneously. Note that AtrousFormer has  two forms, namely global AtrousFormer and local AtrousFormer. The global one is installed on top of the feature extractor to enhance global representation. The local one is used to implement early extracting enhancement, in a local manner. We also propose Local Semantic  Guided  Decoder (LSGD), in which the Gaussian map of the starting point of each lane is employed as the attention map to guide classification process (the distance of each other in this position is long enough). Our contributions can be boiled down into the following points:

\begin{itemize}
	\item We propose global AtrousFormer to collect information in an atrous yet global way, instilling the essence of the ASPP to the transformer structure and finally enhancing the ability to infer lane distribution. 
	\item Global AtrousFormer is evolved into a more compact version, local AtrousFormer, by incorporating slice mechanism. Then it is early embedded into feature extractor like ResNet-18 to enhance feature extraction.
	\item We propose LSGD to obtain more representative feature vector, which then is used to better the lane existence prediction. 
	\item Our network achieves favorable results against other state of the arts on the recent challenging datasets, achieving 78.08 $\rm F1$ score on CULane \cite{SCNN} and 96.71 Accuracy on TuSimple \cite{tusimple}.
\end{itemize}
\section{Related Works}
\label{sec:relatedWorks}
This section will discuss first the deep learning based lane detection methods. In general, the networks to conduct lane detection can be categorized into three classes, semantic segmentation based methods, detection based methods, and key point estimation based methods. Finally, the development of transformer in vision community is sketched.

\textbf{Semantic Segmentation Based Methods.}In this kind of doings, lane detection is modeled as a per-pixel prediction task with an additional classification branch~\cite{rbnet, erfnet, liulane, SCNN, distillation, vanishing, gan, resa}. To improve performance, early they devise various enhancing mechanisms~\cite{rbnet, erfnet, liulane, gan, distillation} and recently resort to feature aggregator modules~\cite{SCNN, resa}.  The \cite{distillation} uses the method of self-distillation to mine spatial properties of lanes. The \cite{erfnet} designs a new powerful backbone for lane detection. The authors in ~\cite{SCNN} use a in-layer, recurrent in four directions, and residual information passing mechanism to enhance the spatial feature representation, finally accommodating the thin and long lane shapes. Later, \cite{resa} mitigates these problems in a sparse gleaning way.

\textbf{Detection Based Methods.} Motivated by the progresses in recent object detection based methods, lane detection research in this direction~\cite{linecnn,laneatt,polylanenet,lstr} can be classified as anchor-based~\cite{linecnn,laneatt} and anchor-free~\cite{polylanenet,lstr}. The \cite{linecnn, laneatt} use beaming lines from either the bottom, left, right to propose primary candidates, and then predicts offset maps to fit the real line shape. However, they are struggling to free themselves from the constraints of anchors. In \cite{laneatt}, local cues are added to refine the problem. The works of \cite{polylanenet,lstr} abandon this kind of doing and directly assume that lanes in the scenes can be modeled as a polynomial equation estimation problem. Although refreshing, they are infeasible to handle the complex scenes, due to overly relying on certain parameters.

\begin{figure*}[htbp]
	\centering
	\includegraphics[scale=0.16]{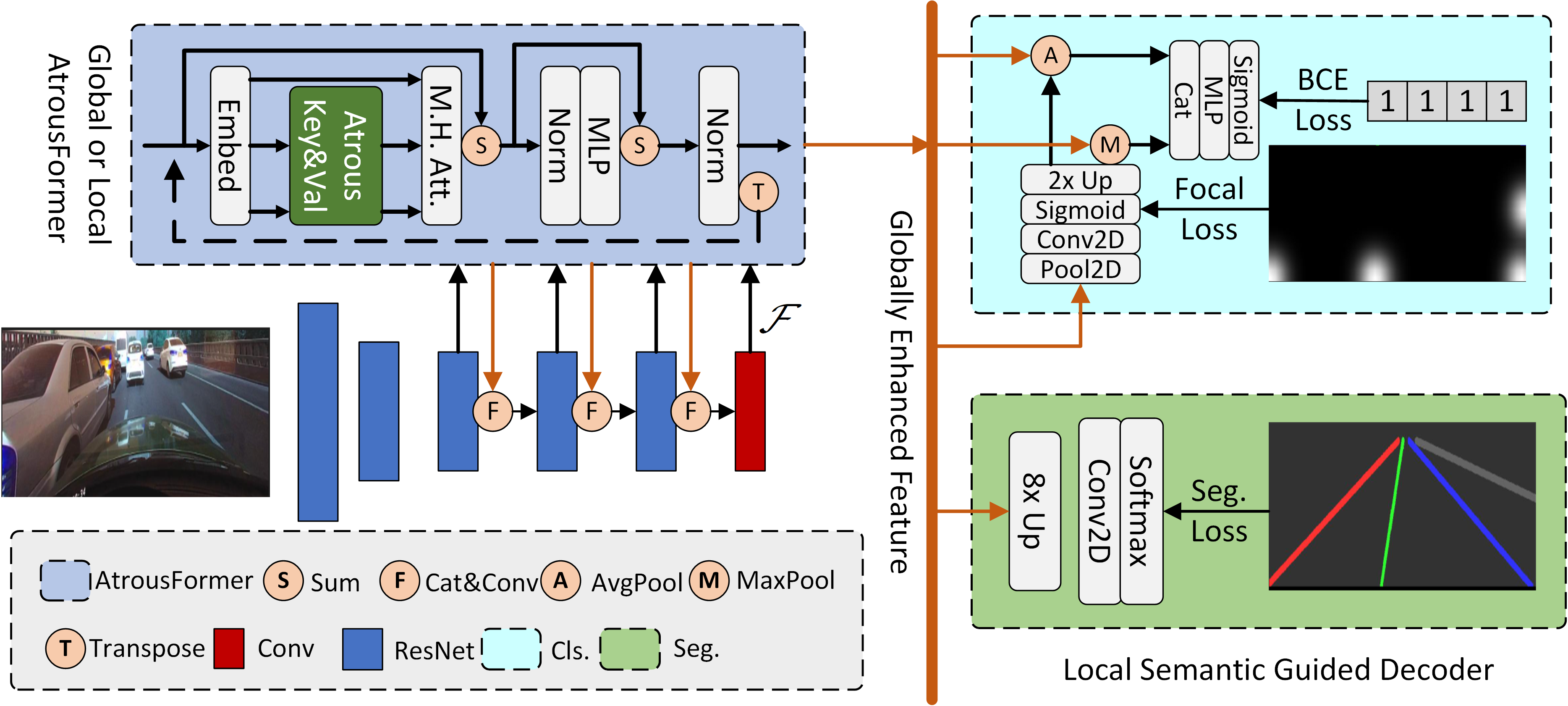}
	\caption{An overview of the architecture. It passes the raw image through local AtrousFormer Enhanced extractor, one $1\times 1$ Convolution, Global AtrousFormer, and Local Semantic Guided Decoder in sequence, finally generating segmentation maps and the corresponding classification scores.}
	\label{arch}
\end{figure*}

\textbf{Key Point Estimation Based Methods.} This approach~\cite{key1,PINet,laneaf,curvelane} usually determines the sparse or dense key points belonging to lanes first via a branch, and simultaneously in other branches predicts their identities to cluster them into different instances. In ~\cite{PINet}, the authors design a branched, multi-task network, including a binary segmentation branch and a identity embedding branch, to clustering key points into different lane instances in an end to end manner. However, in most cases it is suffered from occlusion problem. Therefore, ~\cite{PINet} stack several hourglass modules to enhance the inference ability of the network to alleviate the problem. Recently, \cite{laneaf} proposes to use two affinity matrices to cluster the determined key points to further resolve the problem.

\textbf{Vision Transformers.} The seq2seq Transformer\cite{transformer} is primarily proposed to resolve the problems long haunting around LSTM and various its posteriors, such as long-range information losing, weak semantic representation, and sequential inflexibility. Recently, researchers begin introducing it into vision community to enhance the feature representation of backbone. The Vision Transformer \cite{vit} is the primary one applying  transformer structure on non-overlapping medium-size feature patches for image classification, to some degree achieving speed-accuracy balance.  Later, one of its follow-ups swin transformer \cite{swin} uses window and shifting mechanisms at different scales, further alleviating the overdue computational problem and achieving good performance in different tasks. Others like \cite{pvt, conformer} tinker it in various ways.

\section{Methodology}
\label{sec:methods}
In this section, we show the overall architecture of our network (see Fig.\ref{arch}), the designed global AtrousFormer, local AtrousFormer enhanced extractor and Local Semantic Guided Decoder (LSGD). 

\subsection{Overall Architecture}
We first extract feature using local AtrousFormer enhanced extractor (enhanced ResNet-18, ResNet-34, etc). The max poolings in the third and fourth stages are replaced with dilations. After the raw image passes through the extractor, a feature with spatial size of 1/8 original image is generated. Its channel dimension followingly is compressed to $\rm C^{\it{cmpr}}$ by one $1\times 1$ convolution, denoted as $\mathcal{F}$ of size $\rm H\times W \times C^{\it{cmpr}}$. We then send $\mathcal{F}$ to global AtrousFormer to further enhance the semantic representation, and the enhanced feature is finally fed to the decoder LSGD. The decoder consists of a segmentation branch and a classification branch. In the segmentation branch, the spatially enhanced feature is directly recovered to the original size by $8\times$ bilinear upsampling. And in the sequel, the product is processed by a convolution manipulation to generate the final segmentation maps (the maps later are used to predict the distribution of lane instances). In the classification branch, we primarily get the Gaussian map of the starting point of each lane. Then the map is used to do spatial attention on the enhanced feature. Finally we obtain representative feature vector of each lane to clarify the existence of lanes. In the following context, the global AtrousFormer is first introduced for narrating consistency.

\subsection{Global AtrousFormer}
Let us at the beginning have a retrospect on the seq2seq Transformer. It fuses the thoughts of position-aware embedding, multi-head self- and cross- attention mechanism, and residual connection, making the extracted features more expressive. However, in lane detection, lane itself is slender, and its spreading direction in either the real world or on the image plane follows certain geometric and human-setting rules.  The ignorance of the seq2seq Transformer towards this observation not only causes computational overhead to surge, but also introduces more noises.

The global AtrousFormer imparts the prior knowledge to the seq2seq Transformer, decomposing the sampling process into a two-stage atrous form. The process of the first stage is formulated in the next. The feature $\mathcal{F}$ extracted by the backbone is embedded into three tensors:
\begin{equation} \label{embed}
	\begin{aligned}
		& \mathcal{Q}^{global} = ConvQ_{1\times 1}(\mathcal{F}+\mathcal{P}),\\
		& \mathcal{E}^{key} = ConvK_{1\times 1}(\mathcal{F}+\mathcal{P}),\\
		& \mathcal{E}^{val} = ConvV_{1\times 1}(\mathcal{F}).
	\end{aligned} 
\end{equation}

In \eqref{embed}, $ConvQ_{1\times 1}$ refers to $1 \times 1$ convolution manipulation with stride and padding size equivalent to $1$ and $0$, respectively. Its output $\mathcal{Q}^{global}$ is a tensor of size $\rm{H} \times \rm{W} \times \rm{C}^{\it{cmpr}}$ and can be directly used as the query of the global AtrousFormer. The other two operations $ConvK_{1\times 1}$ and $ConvV_{1\times 1}$ denote $1 \times 1$ convolutions with stride of $1$ and padding size of $\rm{H} \times 0$ (row and column wise paddings). Note that the padding serves only as placeholder that helps the generation of affinity matrices. Both $\mathcal{E}^{key}$ and $\mathcal{E}^{val}$ are of the same size $3\rm{H} \times \rm{W} \times \rm{C}^{\it{cmpr}}$, serving to generate the final key $\mathcal{K}^{global}$ and value $\mathcal{V}^{global}$. The positional encoding tensor $\mathcal{P}$ of size $\rm{H} \times \rm{W} \times \rm{C}^{\it{cmpr}}$ is generated using the same method in \cite{transformer}. 

\begin{figure}[tbp]
	\centering
	\includegraphics[scale=0.24]{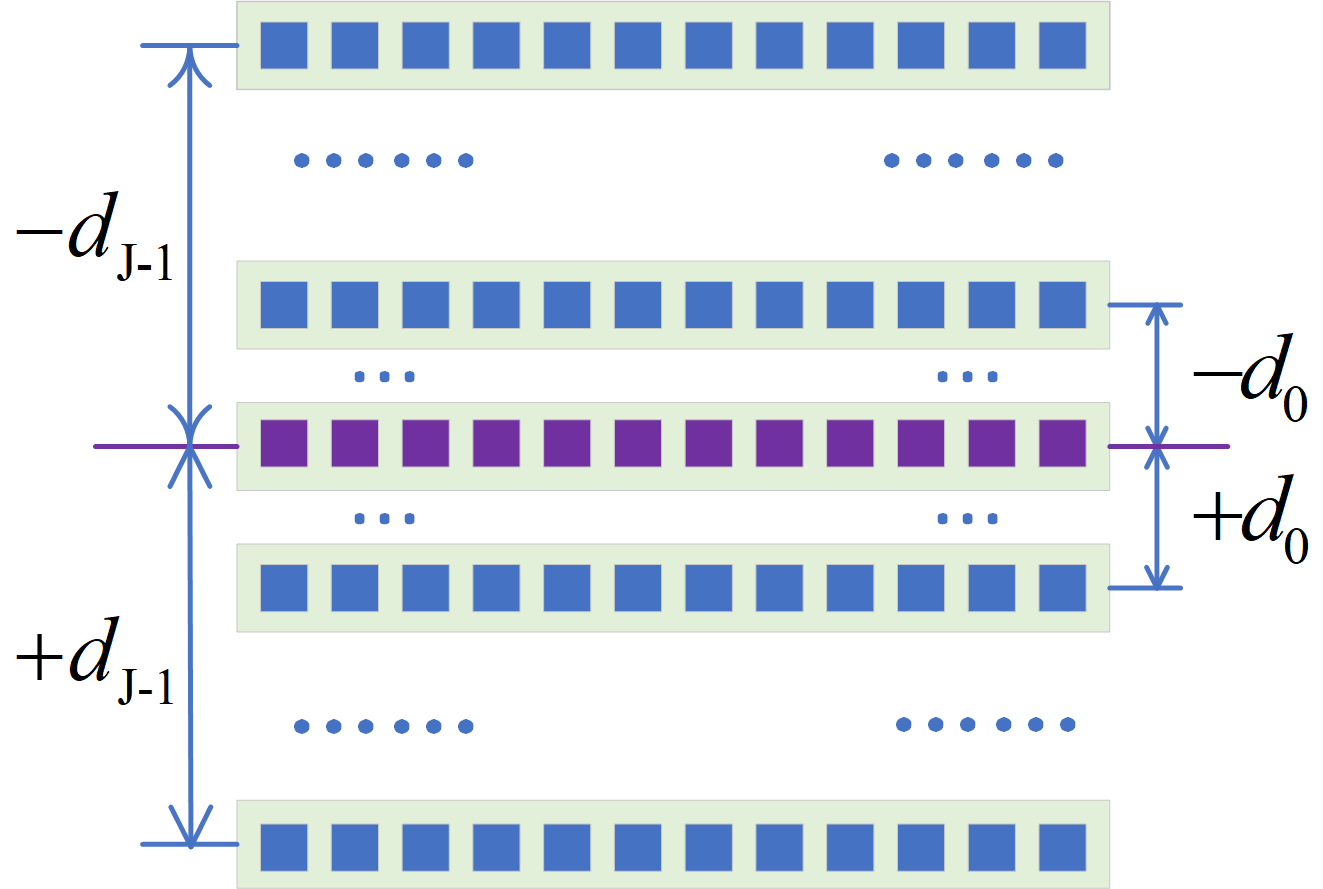}
	\caption{The positions of query entities (in purple boxes) and key\&value entities (in both blue and purple boxes) in global AtrousFormer.}
	\label{fig:att}
\end{figure}

Next we introduce the production of $\mathcal{K}^{global}$ and $\mathcal{V}^{val}$. In global AtrousFormer, members of $\mathcal{Q}^{global}$ in the $i$th row will query information from those that are not only in the $i$th row, but also those above and below the $i$th row in the distances of $\{d_j\}_{j\in \forall}$ (see Fig.\ref{fig:att}):
\begin{equation} \label{heu:exp}
	d_{j} = \lfloor \rm{H}/{2^{(\rm{J}- \it{j})}} \rfloor \rm{,} \hspace{10pt} \it{j} = \rm{0,1,..., \rm{J-1}},
\end{equation}
where $\rm{J}$ is a hyper-parameter specifying the sampling density and $d_{j}\geq 1$. The $\lfloor - \rfloor$ rounds down a float number to its closest integer. The $\mathcal{K}^{global}$ therefore can be generated using the following pattern:
\begin{align}\label{key:gen}
	&\mathcal{K}^{global} = Cat(\mathcal{E}^{key}[\rm{H}: 2\rm{H}, :, :],\nonumber \\
	&\quad \mathcal{E}^{key}[\rm{H}\pm \textcolor{blue}{\it{d}_{\rm 0}}: 2\rm{H}\pm \textcolor{blue}{\it{d}_{\rm 0}}, :, :]\nonumber \\
	&\quad ,..., \nonumber \\
	&\quad \mathcal{E}^{key}[\rm{H}\pm \textcolor{blue}{\it{d}_{\rm J-1}}: 2\rm{H}\pm \textcolor{blue}{\it{d}_{\rm J-1}}, :, :]).
\end{align}

In \eqref{key:gen}, we resort to python conservation word $a:b$ to denote index range from $a$ to $b$ with the default interval of $1$. The $Cat$ concatenates the tensors along the second dimension. The output $\mathcal{K}^{global}$ is of size $\rm{H}\times (2\rm{J}+1)\rm{W} \times \rm{C}^{\it{cmpr}}$. By replacing $\mathcal{E}^{key}$ with $\mathcal{E}^{value}$ in \eqref{key:gen}, we will get $\mathcal{V}^{global}$ of the same size as $\mathcal{K}^{global}$. The heuristic experience is very effective considering the disconnected, thin, and forward-spreading distribution properties of lane instances. More members in the neighboring region are queried than those in the remote area.

Split $\mathcal{Q}^{global}$, $\mathcal{K}^{global}$, and $\mathcal{V}^{global}$ into $\rm{N}^{\it{heads}}$ heads along the channel dimension, each of which has $\rm{C}^{\it{cmpr}}/\rm{N}^{\it{heads}}$ feature maps. As such, the affinity matrices can be computed as follows:
\begin{equation}
	\mathcal{A}^{global} = \frac{Softmax(\mathcal{Q}^{global}@Perm(\mathcal{K}^{global}))}{\sqrt{\rm{\rm{C}^{\it{cmpr}}/N^{\it{heads}}}}},
\end{equation}
in which $Perm$ means to exchange the index order of the second and third dimension, $Softmax$ represents the row-wise normalization, and the symbol $@$ denotes matrix multiplication. The produced affinity matrices $\mathcal{A}^{global}$ is of size $\rm{N^{\it{heads}} \times H \times W \times (2J+1)W}$. Similar to the seq2seq Transformer, we get the final output by using the following manipulations:
\begin{align}
	\label{eq:mlp}
	&\mathcal{I}^{global} = Norm(\mathcal{F} + Recover(\mathcal{A}^{global}@\mathcal{V}^{global}), \nonumber \\
	&\mathcal{F}^{global} = Norm(\mathcal{I}^{global} + MLP(\mathcal{I}^{global})). 
\end{align}

In \eqref{eq:mlp}, $Norm$ and $MLP$ refer to layer normalization and multiple layer perceptrons, respectively. The $Recover$ consisting of consecutive permute and reshape manipulations is used to recover the multi-head feature into the original size. As of now, how members in each row glean information by the proposed atrous way is introduced. The second stage is quite similar to the first stage, first exchanging the row and column indices of $\mathcal{F}^{global}$, then passing it through the equations from \eqref{embed} to \eqref{eq:mlp}, and finally producing the global AtrousFormer enhanced feature.

\subsection{Local AtrousFormer for Early Enhancement}
We further consider interleaving the global AtrousFormer into the early feature extraction stage to enhance the inference ability of our network. However, directly inserting the global AtrousFormer into different stages of ResNet will be costly in terms of hardware and computation overhead. Instead we develop local AtrousFormer by introducing rectangular box restriction to the global AtrousFormer, but in the meanwhile take the spatial distribution attributes of lanes into account. 

We apply the local AtrousFormer to the second, third and fourth stages, all of which generate representative features of size $1/8$ original image (replacing the max poolings in the third and fourth stages with dilated convolution of atrous rates of 2 and 4, respectively). The unenhanced feature after each stage is sent to the local AtrousFormer, and then concatenated with the production to fuse to go to the next stage. This kind of multi-scale fusion objectively strengthens the local information communication, just like the window shifting mechanism in \cite{swin}. 

\begin{figure}[tbp]
	\centering
	\includegraphics[scale=0.24]{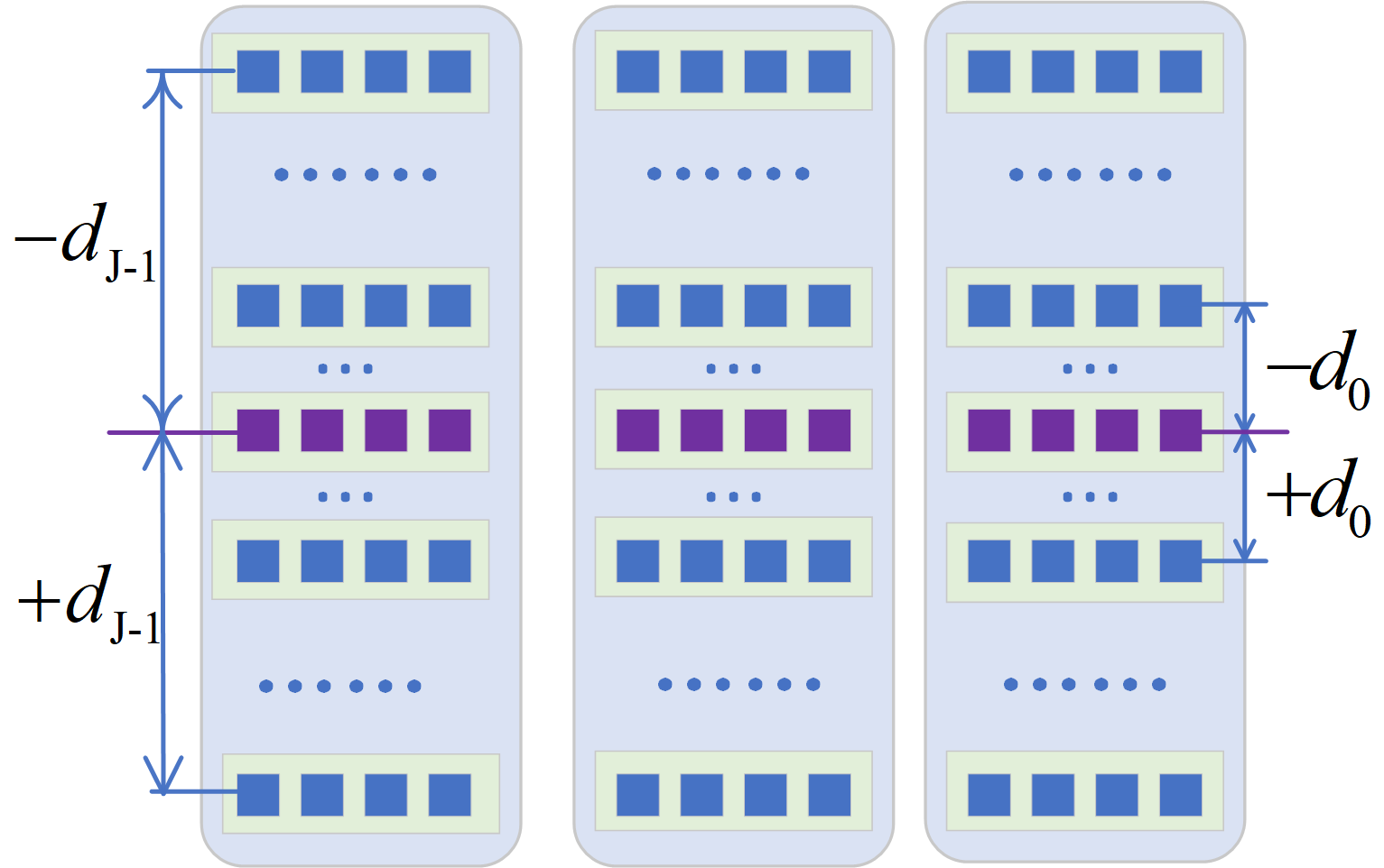}
	\caption{The positions of query entities (in purple boxes) and key\&value entities (in both blue and purple boxes) in local AtrousFormer. The gleaning process happens within each slice.}
	\label{fig:att1}
\end{figure}

Next we give the details of local AtrousFormer in the $i$th stage ($i\in \{2,3,4\}$), which also has gleaning information process of two stages. During the first stage we split the primary feature into $\rm{S}^{\it{col}}$ slices along column, and do the sparse attention only within the slice. After the tensors $\mathcal{Q}_{\it{i}}^{local}$, $\mathcal{E}_{\it{i}}^{key}$, and $\mathcal{E}_{\it{i}}^{val}$ are generated by \eqref{embed}, we transform them into size of $\rm{H}\times \rm{S}^{\it{col}} \times W/\rm{S}^{\it{col}} \times C_{\it{i}}^{\it{stage}}$, $3\rm{H}\times \rm{S}^{\it{col}} \times W/\rm{S}^{\it{col}} \times C_{\it{i}}^{\it{stage}}$, and $3\rm{H}\times \rm{S}^{\it{col}} \times W/\rm{S}^{\it{col}} \times C_{\it{i}}^{\it{stage}}$ using the $Reshape$ operation, respectively. Probing heuristic experience in equation \eqref{heu:exp}  is still adopted in the similar pattern.
The key tensor can be generated (see Fig.\ref{fig:att1}) as follows:
\begin{align}\label{key:gen1}
	&\mathcal{K}_{\it{i}}^{local} = Cat(\mathcal{E}_{\it{i}}^{key}[\rm{H}: 2\rm{H}, :, :, :],\nonumber \\
	&\quad \mathcal{E}_{\it{i}}^{key}[\rm{H}\pm \textcolor{blue}{\it{d}_{\rm 0}}: 2\rm{H}\pm \textcolor{blue}{\it{d}_{\rm 0}}, :, :, :]\nonumber \\
	&\quad ,..., \nonumber \\
	&\quad \mathcal{E}_{\it{i}}^{key}[\rm{H}\pm \textcolor{blue}{\it{d}_{\rm J-1}}: 2\rm{H}\pm \textcolor{blue}{\it{d}_{\rm J-1}}, :, :, :]),
\end{align}
where the operation of $Cat$ will concatenate the tensors along the third dimension, and other manipulations are similar to those previously introduced. The output $\mathcal{K}_{\it{i}}^{local}$ is of size $\rm{H}\times \rm{S}^{\it{col}} \times (2\rm{J}+1))\rm{W}/\rm{S}^{\it{col}}\times C_{\it{i}}^{\it{stage}}$. The value tensor $\mathcal{V}_{\it{i}}^{local}$ of size $\rm{H}\times \rm{S}^{\it{col}} \times (2\rm{J}+1))\rm{W}/\rm{S}^{\it{col}} \times C_{\it{i}}^{\it{stage}}$ also can be produced by replacing $\mathcal{E}_{\it{i}}^{key}$ with $\mathcal{E}_{\it{i}}^{val}$ in \eqref{key:gen1}. 

Split features into $N_{\it{i}}^{\it{heads}}$ heads and the 
affinity matrices for the local AtrousFormer can be computed in the following way:
\begin{equation}
	\mathcal{A}_{\it{i}}^{local} = \frac{Softmax(\mathcal{Q}_{\it{i}}^{local}@Perm(\mathcal{K}_{\it{i}}^{local}))}{\sqrt{\rm{\rm{C_{\it{i}}^{\it{stage}}}/N_{\it{i}}^{\it{heads}}}}},
\end{equation}
in which the $Perm$ manipulation is used to exchange the third and fourth dimension of $\mathcal{K}_{\it{i}}^{local}$. The output $\mathcal{A}_{\it{i}}^{local}$ is of size $\rm{N}_{\it{i}}^{\it{heads}} \times \rm{H}\times \rm{S}^{\it{col}} \times \rm{W/\rm{S}^{\it{col}}}\times (2\rm{J}+1)\rm{W/\rm{S}^{\it{col}}}$. Applying equation \eqref{eq:mlp}, we will get the locally row-wise enhanced feature. The second stage of the local AtrousFormer is first to transpose the row and column indices of input and the rest process is similar to what we have introduced in the end of the last subsection. Compared to attention mechanism in seq2seq Transformer, whose computation of self-attention involves $\rm (HW)^2$ dot products, the computation in global AtrousFormer and local AtrousFormer have $\rm (HW)*(2\rm{J}+1)(H+W)$ and $\rm (HW)*(2\rm{J}+1)(H/S^{\it{row}}+W/S^{\it{col}})$, respectively. The $S^{\it{row}}$  denotes the slice number along row wise.

\subsection{Local Semantic Guided Decoder}
The methods based on semantic segmentation tend to directly apply fully connected layers coupled with $Pooling$ to clarify the existence of each lane in one branch, and $Conv2D$ layers to segment the shapes in the other branch. Both of them are on top of the semantically enhanced feature aggregator. Although the designed feature aggregator is strong enough to handle the segmentation subtask, they ignore the negative influence on the current existence prediction from that of others. Differently from them, we adopt predicting the Gaussian map of the starting point of each lane to guide the classification process, which is named as Local Semantic Guided Decoder (LSGD).

Next we will introduce the details of LSGD. In lane detection, the number of lanes is a prior knowledge specified by the annotation rules, just like $4$ in CULane and $6$ in TuSimple. Therefore, LSGD is able to predict the Gaussian map of the starting point of each lane as follows:
\begin{equation}
	\mathcal{G}^{map} = Sig((Conv(MaxPool_{2d}(\mathcal{F}^{global})))),
\end{equation}
in which $Conv$ of consists of two consecutive $1$ stride convolutions with kernels of size $3\times 3 \times \rm{C}^{\it{cmpr}} \times \rm{C}^{\it{cmpr}}$ and $1\times 1 \times \rm{C}^{\it{cmpr}} \times \rm{N}^{\it{lanes}}$ (number of lanes), respectively, ${MaxPool}_{2d}$ represents $2\times$ maxpooling, and $Sig$ represents sigmoid manipulation. The output of size $\rm N^{\it{lanes}}\times H/2 \times W/2$ is supervised by the focal loss in \cite{cornernet}. The $i$th Gaussian map $\mathcal{G}_{[i,:,:]}^{map}$ serves as the attention map to 
guide the classification process of the $i$th lane:
\begin{align}\label{attention}
	c^{score}_{i} &= Sig(MLP(Cat(MaxPool(\mathcal{F}^{attended}_{i}), \nonumber\\
	&\qquad \qquad \qquad  AvgPool(\mathcal{F}^{attended}_{i})))),
\end{align}
in which $\mathcal{F}^{attended}_{i} = Up_2(\mathcal{G}_{[i,:,:]}^{map}) \otimes \mathcal{F}^{global}$, where $Up_2$ is bilinear interpolation of $2\times$ upsampling; $MaxPool$ and $AvgPool$ refer to the max and average poolings, respectively; the $MLP$ is three fully connected layers with hidden dimension as $64$ and $16$. The segmentation maps can be obtained by passing $\mathcal{F}^{global}$ through a $8\times$ bilinear interpolation and a convolution manipulation with stride $1$ and kernel size $3\times 3 \times \rm{C}^{\it{cmpr}} \times \rm{N}^{\it{lanes}}$.

\section{Experiments}
\subsection{Datasets}
To verify the effectiveness of our proposed methods, we demonstrate their respective performance on three currently wide-used datasets, CULane~\cite{SCNN}, TuSimple~\cite{tusimple}, and BDD100~\cite{bdd}. They are challenging and encompass a variety of scenarios. 

To be specific, CULane has 55 hours video clips, and consists of nine road conditions:  normal, crowd, curve, dazzle night, night, no line, and arrow. Most of the frames within are shot in the urban, and a few are in the rural or highway. In total, CULane has $133235$ frames of size $590\times 1640$, of which 88880 are used for training, 9675 are used to validate, and the left are used to benchmark the performance of the model. The number of lanes in each frame is at most $4$.

The samples in TuSimple are shot under stable light situation by vehicles in highways. Compared to CULane, the dataset is relatively small. In total, it has $6408$ frames. Of them, 3236 are used for training, 358 are used for validation, and 2782 are used to test. All are of size $720\times 1280$, and each frame contains at most $5$ lanes. 

The BDD100K originally designed for lane classification has binary annotation for lane semantics (provided by \cite{distillation}). The lanes are close to each other, and also embody in different forms, thus challenging to current algorithms. Different from CULane and TuSimple, the width of each lane in the training set is set to 8, while in the testing set it is set to 2. Following \cite{distillation}, we use the training set of 80000 frames of resolution $720\times 1680$ to train our model, and validation set of 10000 frames of the same resolution to test.

\subsection{Implementation Details}
We resize the original images to $288\times 800$ for CULane and $368\times 640$ for TuSimple, respectively. Furthermore, our training involves the following techniques of data augmentation: random scaling, cropping, random rotation, color jittering, and etc.
The optimizer uses $SGD$ with momentum 0.9 and weight decay 1e-4 to train our model. The learning rate is set to 2.5e-2 for CULane and 2.0e-2 for TuSimple, respectively. Warming-up strategy is used in the first 500 batches. Polynomial learning rate decay policy with power set to 0.9. The loss function is the summation of the focal loss multiplied by $0.2$ and the one in \cite{SCNN,resa}, which consists of segmentation $BCE$ loss and existence classification $BCE$ loss. The batch size is set to $8$ for CULane and $4$ for TuSimple. The number of training epoch is set to $12$ for CULane and 80 for TuSimple. All models are trained on 3 NVIDIA 2080Ti GPUs with Pytorch framework. For local AtrousFormer enhanced backbone, we set $\rm{N}^{heads}$ in the second, third, and fourth stages to $2$, $8$, and $16$, respectively, let $\rm{J}$ equal to $4$, and set slice size to $(18,20)$ and $(23, 20)$ on CULane and TuSimple, respectively. The feature extracted from the backbone are compressed to $128$ using $1\times 1$ convolution. For global AtrousFormer, we set $\rm{J}$ to 4 and $\rm{N}^{\it{heads}}$ to $16$. The threshold values are set to $0.5$ in both the classification branch and segmentation branch. The settings for BDD100K is slightly different from what we have adpted in CULane. Following \cite{distillation}, we resize the training image to $360\times 720$ in training and use the unresized image to test. During the experiments on BDD100K, the branch of local semantic guidance is removed.

For CULane, lanes are treated as a line of 30-pixel-width. The intersection-over-union (IOU) of the predictions and groundtruth lanes is used. In general, the predicted lanes that have a IOU with their corresponding targets larger or equivalent to the threshold value 0.5 are defined as the true positives (TP). The FP and FN are used to denote false positives and false negatives, respectively. The $\rm F1$-measure is adopted by CULane to evaluate the results, as follows:
\begin{equation}
	\rm{F}_{1} = \frac{2\times Precision \times Recall}{Precision + Recall},
\end{equation}
where $\rm Precision=TP/(TP+FP)$ and $\rm Recall=TP/(TP+FN)$.

The TuSimple evaluates the results by Accuracy, defined as follows:
\begin{equation}
	\rm Accuracy = \sum\nolimits_{\it{clip}}\frac{C^{\it{clip}}}{S^{\it{clip}}},
\end{equation}
in which $\rm C^{\it{clip}}$ represents the number of correctly predicted lane points. To be specific, they are assigned the right identities. Their distances with the corresponding ground truth points are within certain range in a clip. The $\rm S^{\it{clip}}$ denotes the number of ground truth points in a clip.

For BDD100K, following \cite{distillation}, we use per-pixel accuracy and mean IOU to evaluate the performance of different models.
\begin{table*}[htbp]
	\centering
	\caption{Quantitative results of our network versus other state of the arts under nine situations of CULane test set against the $\rm F1$ metric. The Cross only reports FP. Note that red, blue, and green colors represents the highest, second, and third scores, respectively. The symbol of '-' represents the results are not reported officially. }
	\resizebox{\textwidth}{!}
	{
		\setlength\tabcolsep{1pt}
		\renewcommand\arraystretch{1.6}
		\begin{tabular}{l||ccccccccc|c|c}
			\toprule[1pt]
			Methods  &Normal $\uparrow$ &Crowded $\uparrow$ &Dazzle $\uparrow$ &Shadow$\uparrow$ &NoLine $\uparrow$ &Arrow$\uparrow$ &Curve$\uparrow$ &Cross$\downarrow$ &Night$\uparrow$ &Total$\uparrow$ &Speed (ms)\\
			\hline
			\hline
			Fast Draw~\cite{fastdraw} &85.90  &63.60 &57.00 &59.90 &40.60 &79.40 &65.20 &7013 &57.80 &- &11 \\
			SpinNet~\cite{spinnet} &90.50 &71.70 &62.00 &72.90 &43.20 &85.00 &50.70 &-  &68.10 &74.20 &-\\
			R18-UFAST~\cite{ufast}&87.70 &66.00 &58.40 &62.80 &40.20 &81.00 &57.90 &1743 &62.10 &68.40 &\textcolor{red}{\bf{3}}\\
			R34-UFAST~\cite{ufast}&90.70 &70.20 &59.50 &69.30 &44.40 &85.70 &69.50 &2037 &66.70 &72.30 &6\\
			R18-E2E~\cite{e2e}&90.00 &69.70 &60.20 &62.50 &43.20 &83.20 &70.30 &2296 &63.30 &70.80 &-\\
			R34-E2E~\cite{e2e} &90.40 &69.90 &61.50 &68.10 &45.00 &83.70 &69.80 &2077 &63.20 &71.50 &-\\
			ERFNet-E2E~\cite{e2e} &91.00 &73.10 &64.50 &74.10 &46.60 &85.80 &71.90 & 2022 &67.90 &74.00&-\\
			R34-SCNN~\cite{SCNN} &90.60 &69.70 &58.50 &66.90 &43.40 &84.10 &64.40 &1990 &66.10 &71.60 &116\\
			ERFNet-SAD~\cite{distillation} &90.10 &68.80 &60.20 &65.90 &41.60 &84.00 &65.70 &1998 &66.00 &70.80 &10\\ 
			ERFNet-IntRA-KD &- &- &- &- &- &- &- &- &- &72.40 &-\\
			PINet~\cite{PINet} &90.30 &72.30 &66.30 &68.40 &49.8 &83.70 &65.60 &1427 &67.70 &74.40 &40\\
			SIM-CycleGAN\cite{liulane} &91.80 &71.80 &66.40 &76.20 &46.10 &87.80 &67.10 &2346 &69.40 &73.90 &-\\
			CurveLanes-NAS-S~\cite{curvelane} &88.30 &68.60 &63.20 &68.00 &47.90 &82.50 &66.00 &2817 &66.20 &71.40 &-\\
			CurveLanes-NAS-M~\cite{curvelane} &90.20 &70.50 &65.90 &69.30 &48.80 &85.70 &67.50 &2359 &68.20 &73.50&-\\
		    R34-RESA~\cite{resa} &91.90 &72.40  &66.50 &72.00 &46.30 &88.10 &68.60 &1896 &69.80 &74.50 &22 \\
			R18-LaneATT~\cite{laneatt} &91.11 &72.96 &65.72 &70.91 &48.35 &85.49 &63.37 &1170 &68.95 &75.09 &\textcolor{blue}{\bf{4}}\\
			R34-LaneATT~\cite{laneatt} &92.14 &75.03 &66.47 &\textcolor{red}{\bf{78.15}} &49.39 &\textcolor{green}{\bf{88.38}} &67.72 &1330 &70.72 &76.68 &\textcolor{green}{\bf{6}}\\
			R122-LaneATT~\cite{laneatt} &91.74 &\textcolor{blue}{\bf{76.16}} &\textcolor{blue}{\bf{69.47}} &\textcolor{green}{\bf{76.31}} &\textcolor{red}{\bf{50.46}} &86.29 &68.40 &1746 &68.90 &77.02 &45\\
			\hline\hline
			R18-Light &\textcolor{blue}{\bf{92.77}} &74.69 &66.89 &69.68 &49.25 &88.09 &\textcolor{blue}{\bf{70.59}} &\textcolor{blue}{\bf{1096}} &\textcolor{green}{\bf{72.85}} &\textcolor{green}{\bf{77.03}} &20\\
			
			XR18-Ours &\textcolor{green}{\bf{92.72}} &\textcolor{green}{\bf{75.56}} &\textcolor{green}{\bf{68.16}} &73.67 &\textcolor{green}{\bf{50.07}} &\textcolor{red}{\bf{88.82}} &\textcolor{green}{\bf{70.32}} &\textcolor{green}{\bf{1169}} &\textcolor{blue}{\bf{73.49}} &\textcolor{blue}{\bf{77.63}} &36\\
			
			XR34-Ours &\textcolor{red}{\bf{92.83}} &\textcolor{blue}{\bf{75.96}} &\textcolor{red}{\bf{69.48}} &\textcolor{blue}{\bf{77.86}} &\textcolor{blue}{\bf{50.15}} &\textcolor{blue}{\bf{88.66}} &\textcolor{red}{\bf{71.14}} &\textcolor{red}{\bf{1054}} &\textcolor{red}{\bf{73.74}} &\textcolor{red}{\bf{78.08}} &44\\
			\bottomrule[1pt]  
		\end{tabular}
	}
	\label{table:culane}
\end{table*}

\begin{table}
	\centering
	\caption{Quantitative comparison results with other state of the arts on the test set of TuSimple in terms of Accuracy, FP, and FN metrics.}
	\resizebox{0.55\textwidth}{!}{
		\setlength\tabcolsep{8pt}
		\renewcommand\arraystretch{1}
		\begin{tabular}{l||c|c|c}
			\hline
			Methods &Accuracy $\uparrow$ &FP$\downarrow$ &FN$\downarrow$\\
			\hline
			\hline
			SCNN~\cite{SCNN}  &96.53 &6.17 &\textcolor{red}{\bf{1.80}}\\
			EL-GAN~\cite{gan}  &94.90 &4.12 &3.36 \\
			PINet~\cite{PINet}  &\textcolor{green}{\bf{96.70}} &2.94 &\textcolor{blue}{\bf{2.63}} \\
			ENet-SAD~\cite{distillation}&96.64 &6.02 &\textcolor{green}{\bf{2.05}} \\
			ERF-E2E~\cite{e2e}  &96.02 &3.21 &4.28 \\
			FastDraw~\cite{fastdraw}  &95.20 &7.60 &4.50 \\
			R18-UFAST~\cite{ufast}  &95.82 &19.05 &3.92 \\
			R34-UFAST~\cite{ufast} &95.86 &18.91 &3.75 \\
			PolyLaneNet~\cite{polylanenet}  &93.36 &9.42 & 9.33 \\
			LSTR~\cite{lstr}  &96.18 &\textcolor{green}{2.91} & 3.38 \\
			R18-RESA~\cite{resa}  &\textcolor{red}{\bf{96.82}} &3.95 & 2.83 \\
			R34-RESA~\cite{resa}  &\textcolor{green}{\bf{96.70}} &3.96 & 2.48 \\
			R18-LaneATT~\cite{laneatt}  &95.57 &3.56 &3.01 \\
			R34-LaneATT~\cite{laneatt}  &95.63 &3.53 &2.92 \\
			\hline
			\hline
			XR18-Ours  &96.59 &\textcolor{blue}{\bf{2.83}} & 3.26 \\
			XR34-Ours  &\textcolor{blue}{\bf{96.71}} &\textcolor{red}{\bf{2.82}} & 3.24 \\
			\hline
	\end{tabular}}	
	\label{tab:tusimple}
\end{table}

\begin{table}
	\centering
	\caption{Quantitative comparison results with other state of the arts on the test set of BDD100K in terms of average per-pexel accuracy and IOU of lane semantic of each frame.}
	\resizebox{0.55\textwidth}{!}{
		\setlength\tabcolsep{8pt}
		\renewcommand\arraystretch{1}
		\begin{tabular}{l||c|c}
			\hline
			Methods &Accuracy $\uparrow$ &IOU$\uparrow$\\
			\hline
			ResNet18~\cite{resnet} &30.66 &11.07\\
			ResNet34~\cite{resnet} &30.92 &12.24\\
			SCNN~\cite{SCNN}  &35.79 &15.04\\
			R18-SAD~\cite{distillation} &31.10 &13.29 \\
			R34-SAD~\cite{distillation} &32.68 &14.56 \\
			ERFNet-SAD~\cite{distillation} &\textcolor{green}{\bf{36.56}} &\textcolor{green}{\bf{16.02}}\\
			\hline
			\hline
			XR18-Ours  &\textcolor{blue}{\bf{59.09}} &\textcolor{blue}{\bf{22.81}}  \\
			XR34-Ours  &\textcolor{red}{\bf{59.20}} &\textcolor{red}{\bf{23.31}} \\
			\hline
	\end{tabular}}	
	\label{tab:bdd}
\end{table}

\subsection{Comparison with the SOTAs}
\begin{figure*}[tbp]
	\includegraphics[width=\textwidth]{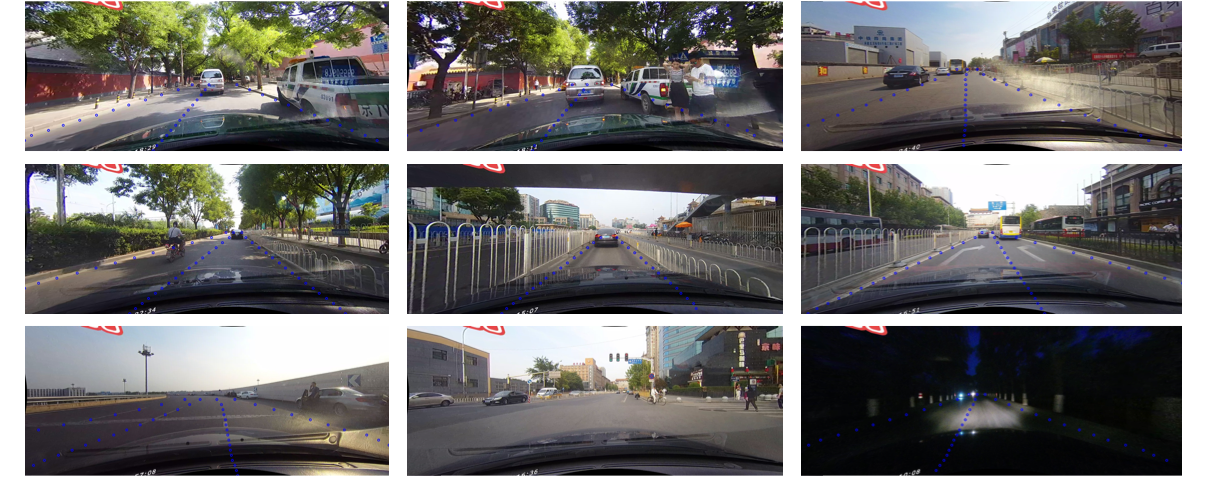}\\ 
	\centering
	\caption{Visualization results of R34-Ours under nine scenarios of CULane,  namely, Normal, Crowded, Dazzle, Shadow, NoLine, Arrow, Curve, Cross and Night (from left to right, from top to bottom, best viewed by zooming in). Note that in the Cross situation, only FP is calculated.} 		
	\label{fig:vis}
\end{figure*}

In this section, we will demonstrate the performance of our network on CULane, TuSimple, and BDD100K, respectively. For CULane, the quantitative and visualization results under nine scenarios are showed in Tab.\ref{table:culane} and Fig.\ref{fig:vis}, respectively.
The quantitative results for TuSimple and BDD100K are listed in Tab.\ref{tab:tusimple} and Tab.\ref{tab:bdd}. Algorithms to compare include FastDraw~\cite{fastdraw}, SpinNet~\cite{spinnet}, UFAST~\cite{ufast}, R18-E2E~\cite{e2e}, R34-E2E~\cite{e2e}, ERFNet-E2E~\cite{e2e}, SCNN~\cite{SCNN}, ENet-SAD~\cite{distillation}, ERFNet-IntRA-KD~\cite{inter}, PINet~\cite{PINet},  Curvelanes-NAS~\cite{curvelane}, RESA~\cite{resa}, LaneATT~\cite{laneatt}, and SIM-CycleGan~\cite{liulane}. All of the results are borrowed from their official publishings. 

$\textbf{CULane Results Analysis.}$ In Tab.\ref{table:culane}, the XR18-Ours and XR34-Ours are used to represent implementations based on enhanced ResNet-18 and ResNet-34, respectively. The R18-Light represents the one of ResNet18+Local AtrousFormer (slice size (18, 20))+LSGD. It can be seen that overall our network achieves 70.03, 77.63, and 78.08 scores in terms of $\rm{F1}$ metric. Our encoder-enhanced network (XR18-Ours and XR34-Ours) surpasses the current state of the art LaneATT by 2.54 and 1.4 points on ResNet-18 and ResNet-34, respectively. As for those adopting temporally-recurrent based segmentation methods on ResNet34, SCNN and RESA, XR34-Ours tops them by 3.58 and 6.48 points, respectively. For different environments of Normal, Crowded, Dazzle, Shadow, NoLine, Arrow, Curve, Cross and Night, 
XR34-Ours achieves (1) 0.69, 0.93, 3.01, -0.29, 0.76, 0.28, 3.42, 276 and 3.02 gains more than LaneATT, (2) 2.23, 6.26, 10.98, 10.96, 6.75, 4.56, 6.74,936, and 7.64 gains more than SCNN, and (3) 0.93, 6.62, 2.98, 5.86, 3.85, 0.56, 2.54, 842, and 3.94 gains more than RESA on ResNet34.  Coupled with the visualization results of R34-Ours in Fig.\ref{fig:vis}, we can safely say that our network has a strong inference ability. It is able to complete the lane instance according to the environment, even in the dark situation (see the ninth image).
In terms of running time, R18-Light tops SCNN and RESA by 96 ms and 2 ms, respectively, while keeping satisfying performance. The XR34-Ours is approximately $3\times$ faster than our predecessor SCNN on ResNet34.

$\textbf{TuSimple Results Analysis.}$ The quantitative results on TuSimple in Tab.\ref{tab:tusimple} to validate the performance of our network. It can be seen that our network obtains an accuracy of 96.59 and 96.71 on ResNet-18 and ResNet-34, respectively. The results are very approximate to the highest scores achieved by RESA. Under metric FP, our method gets the best performance of 2.82.

$\textbf{BDD100K Results Analysis.}$ On BDD100k, only some of the semantic segmentation methods can be reported due to the constraints of complex lane forms in Tab.\ref{tab:bdd}.  This is also one of the most important merits of our AtrousFormer, able to model more complex lanes. The results are showed in Tab\ref{tab:bdd}. It can be seen that XR18-Ours surpasses ERFNet-SAD by 22.53 points and 6.79 points in terms of accuracy and IOU metrics, respectively.

\subsection{Ablation Studies}
$\textbf{Effectiveness of Main Contributions.}$ To validate the effectiveness of the local AtrousFormer enhanced encoder, the global AtrousFormer, and LSGD, we provide detailed ablation studies on CULane in Tab.\ref{tab:ablation}. We stipulate ResNet-18+LSGD (without local attention guidance) as the baseline. The segmentation maps and their corresponding existence scores can be obtained after passing the parallel branches of $8\times$ $UpSampling$\&$Conv2D$ and $8\times$ $Pooling$\&$MLP$. In Tab.\ref{tab:ablation}, the A.F. and XR are used to represent AtrousFormer and locally enhanced ResNet, respectively. We next give some analysis on the adopted techniques.

\begin{table}
	\centering
	\caption{Ablation studies on the test set of CULane against $\rm F1$, Precision, and Recall metrics.}
	\resizebox{0.6\textwidth}{!}{
		\setlength\tabcolsep{3pt}
		\renewcommand\arraystretch{1.0}
		\begin{tabular}{l|c|c|c}
			\hline
			Methods &F1$\uparrow$  &Precision$\uparrow$ &Recall$\uparrow$\\
			\hline
			\hline
			Baseline  &69.60 &70.02 &69.60\\
			+Global A.F.  &75.17 &76.66 &73.74\\
			+Global A.F.+LSGD  &77.19 &82.56 &72.49\\
			XR18+global A.F.+LSGD  &77.63 &83.73 &72.36\\
			XR34+global A.F.+LSGD  &78.08&84.36&72.67\\
			\hline
	\end{tabular}}	
	\label{tab:ablation}
\end{table}

\begin{table}[t]
	\centering
	\caption{The impact of slice size on XR18-Ours}.
	\resizebox{0.6\textwidth}{!}{
		\setlength\tabcolsep{2pt}
		\renewcommand\arraystretch{1.2}
		\begin{tabular}{l|c|c|c|c}
			\hline
			Slice Size &(36, 100) &(18, 20) &(9, 10) &(3, 5)\\
			\hline
			F1 (R18-Ours)  &- &\textcolor{red}{77.63}  &77.56 &56.10\\
			\hline
	\end{tabular}}	
	\label{tab:ss}
\end{table}
\begin{table}[t]
	\centering
	\caption{The impact of experience $\rm{J}$ on XR18-Ours}.
	\resizebox{0.6\textwidth}{!}{
		\setlength\tabcolsep{4pt}
		\renewcommand\arraystretch{1.4}
		\begin{tabular}{l|c|c|c|c|c|c}
			\hline
			Experience $\rm{J}$ &5 &4  &3 &2 &1 &0\\
			\hline
			F1 (R18-Ours)  &77.61  &\textcolor{red}{77.63} &77.56 &76.42 &76.19 &74.68\\
			\hline
	\end{tabular}}	
	\label{tab:experience}
\end{table}
\begin{table}[t]
	\centering
	\caption{Comparison results of the seq2seq Transformer and the global AtrousFormer along the number of heads on ResNet-18+LSGD (The symbol of '-' denotes running out of memory.)}.
	\resizebox{0.6\textwidth}{!}{
		\setlength\tabcolsep{2pt}
		\renewcommand\arraystretch{1.2}
		\begin{tabular}{l|c|c|c|c}
			\hline
			Methods (Heads) &F1$\uparrow$  &Precision$\uparrow$ &Recall$\uparrow$ &K\&V Pos. $\downarrow$\\
			\hline
			\hline
			Seq2seq Trm. (1)  &76.19 &82.19 &71.01 &3600\\
			Global A.F. (1)  &76.44 &82.24 &71.41 &1224\\
			Seq2seq Trm. (4)  &76.37 &82.12 &71.38 &14400\\
			Global A.F. (4)  &76.84 &82.29 &72.05 &4896\\
			Seq2seq Trm. (8)  &76.42 &81.89 &71.65 &28800\\
			Global A.F. (8)  &76.85 &82.29 &72.08 &9792\\
			Seq2seq Trm. (16)  &- &- &- &57600\\
			Global A.F. (16)& 77.19 &82.56 &72.49 &19584\\
			\hline
			\hline
			R.G. A.F. (16) &76.84 &82.30 &72.07 &14400\\
			C.G. A.F. (16) &76.27 &82.97 &70.57 &5184 \\
			\hline
	\end{tabular}}	
	\label{tab:transformer}
\end{table}
It can be seen that AtrousFormer is very effective on improving the performance of lane detection. Compared to baseline, equiping global AtrousFormer results in 5.57 points of F1 gains on ResNet-18. The designed decoder LSGD coupled with global AtrousFormer surpasses the global AtrousFormer+LSGD without attention mechanism by 2.02 points in terms of $\rm F1$ metric, which proves the effectiveness of the local attention guidance. Improving feature extractor from ResNet-18 to ResNet-34 will generate 0.48 point improvement. 

$\textbf{Selectrions of Atrous Experience and Slice Size.}$ To avoid the laboriousness of super-parameter searching, we at first accept the in-layer fusion experience in \cite{resa} to bootstrap, namely $\rm{J}=4$, tuning XR18-Ours in a greedy manner. Then we study the impact of slice size of local AtrousFormer on XR18-Ours and report the results in Tab.\ref{tab:ss}. It turns out that when slice size is set to (18, 20), the network gets the best performance. The results confirm that probing lane distribution needs big enough boxes. Then turning up or down $\rm{J}$, we find that when $\rm{J}=4$, our network gets the best performance. Along the decreasing of sampling density, the performance of network deteriorates significantly. Related results is listed in Tab.\ref{tab:experience}.

$\textbf{Compared to the Seq2Seq Transformer.}$ In Tab.\ref{tab:transformer}, we provide comparison results of the global AtrousFormer and the seq2seq Transformer. All are conducted on ResNet-18+LSGD by controlling the number of heads. It can be seen that compared to the seq2seq Transformer, in terms of the computation of affinity matrices and F1, the AtrousFormer is more efficient than the seq2seq Transformer given the sparsity of its queried positions. The number comparison is also listed (abbreviated as $\rm K\&V$ Pos. in the last column). 

$\textbf{Effectiveness of the Two-stage Strategy.}$ We validate the effectiveness of the two-stage strategy in the last two rows in Tab.\ref{tab:transformer}. The symbols of R.G. and C.G. represent only row-wise and only column-wise global AtrousFormers, respectively. Individually using them will achieve 76.84 and 76.27 points of the $\rm F1$ score, respectively. The fully global AtrousFormer surpasses the only row-wise one and only column-wise one by 0.35 and 0.98 point, respectively.

\section{Conclusion}
In this paper, we design AtrousFormer to enhance network inference ability and mine the spatial distribution of lanes. The proposed AtrousFormer mainly has two representative forms, i.e., global AtrousFormer and local AtrousFormer, both of which glean information first by rows and then by columns according to the specified heuristic experience. In this paper, the former is installed on top of the extractor to enhance the global representation of the extracted feature. The latter one has a more compact form, and is inserted into the feature extractor to strengthen extraction. In addition, we design local semantic guided decoder to incorporate the local information into classification stage. Our network achieves impressive performance on CULane, TuSimple and BDD100K. It is able to handle complex situations like occullusion by other vehicles, bad weather conditions, abrasion, terrible illumination intensity, and compleax lane forms.  

%



\bibliographystyle{elsarticle-num}

\bibliography{sample}

\end{document}